\documentclass[draft]{agujournal2019}

\draftfalse

\journalname{Journal of Advances in Modeling Earth Systems (JAMES)}

\usepackage{amsmath,amsfonts,amssymb,amsthm}
\usepackage{cleveref}
\usepackage{mathtools}

\usepackage{url} %this package should fix any errors with URLs in refs.
\usepackage{soul}

\newcommand{\bs}{\boldsymbol}
\newcommand{\bb}{\mathbb}
\newcommand{\mrm}{\mathrm}
\newcommand{\mcl}{\mathcal}

\newcommand{\bx}{ \bs x}
\newcommand{\by}{ \bs y}
\newcommand{\PX}{ \bb P_{\bb X} }
\DeclareMathSymbol{\shortminus}{\mathbin}{AMSa}{"39}

\newcommand{\revs}[1]{\textcolor{black}{#1}}
\newcommand{\nrevs}[1]{\textcolor{black}{#1}}

\newtheorem{definition}{Definition}
\newtheorem{theorem}{Theorem}
\newtheorem*{theorem*}{Theorem}

\begin{document}

\title{Data Assimilation Networks}

\authors{Pierre Boudier\affil{1}, Anthony Fillion\affil{2}, Serge Gratton\affil{2}, Selime G\"urol\affil{3}, Sixin Zhang\affil{2}}

\affiliation{1}{NVIDIA / ANITI, Toulouse, France}
\affiliation{2}{Universit\'e de Toulouse / ANITI, Toulouse, France}
\affiliation{3}{CERFACS / ANITI, Toulouse, France}
%\affiliation{=number=}{=Affiliation Address=}
%(repeat as many times as is necessary)

%% Corresponding Author:
% Corresponding author mailing address and e-mail address:

% (include name and email addresses of the corresponding author.  More
% than one corresponding author is allowed in this LaTeX file and for
% publication; but only one corresponding author is allowed in our
% editorial system.)

% Example: \correspondingauthor{First and Last Name}{email@address.edu}

\correspondingauthor{Sixin Zhang}{sixin.zhang@irit.fr}

%% Keypoints, final entry on title page.

%  List up to three key points (at least one is required)
%  Key Points summarize the main points and conclusions of the article
%  Each must be 140 characters or fewer with no special characters or punctuation and must be complete sentences

% Example:
% \begin{keypoints}
% \item	List up to three key points (at least one is required)
% \item	Key Points summarize the main points and conclusions of the article
% \item	Each must be 140 characters or fewer with no special characters or punctuation and must be complete sentences
% \end{keypoints}

\begin{keypoints}
\item We propose a general framework \revs{Data Assimilation Networks (DAN)} based on an extended Elman Network 
for Bayesian Data Assimilation. 
\item We show that DAN can achieve optimal prior and posterior density estimations by optimizing likelihood-based objective functions.
\item Numerically DAN achieve comparable
performance to EnKF methods on Lorenz-95 system,
without using explicit regularization such as localization or inflation.

\end{keypoints}

% ABSTRACT

\begin{abstract}
\justify
Data Assimilation (DA) aims at estimating the posterior conditional 
probability density functions based on error statistics of the noisy
observations and the dynamical system.
State of the art methods are sub-optimal due to 
the common use of Gaussian error statistics and the linearization of the non-linear dynamics. 
To achieve a good performance, these methods often require case-by-case fine-tuning 
by using explicit regularization techniques such as inflation and localization.
In this paper, we propose a \textit{fully data driven deep learning framework}
generalizing recurrent Elman networks and data assimilation algorithms. 
Our approach approximates a sequence of prior and posterior densities conditioned on noisy observations using a \textit{log-likelihood cost function}.
By construction our approach can then be used for general nonlinear dynamics 
and non-Gaussian densities. 
As a first step, we evaluate the performance of the proposed approach by using fully and partially observed Lorenz-95 system in which the outputs of the recurrent network are fitted to Gaussian densities.
We numerically show that our approach, \textit{without using any explicit regularization technique}, 
achieves comparable performance to the state-of-the-art methods, IEnKF-Q and LETKF, across various ensemble size.
\end{abstract}

\section*{Plain Language Summary}
\justify
Data Assimilation (DA) aims at forecasting the state of a dynamical system by combining information coming from the dynamics and noisy observations. Bayesian data assimilation uses the random nature of a system to predict its states in terms of probability density functions (pdfs). 
The calculation of these densities is difficult for non-linear dynamical systems. 
Practical algorithms compute limited statistics due to computational cost,
but this results in sub-optimal DA algorithms which requires then the use of explicit regularization techniques to increase the performance of the algorithm.

With the advances in Machine Learning (ML) and deep learning, 
there has been significant increase in the research of using 
ML for data assimilation to decrease the 
computational cost, or to have better estimation of the state. 
In this paper, we propose a fully data driven algorithm to learn 
the prior and posterior pdfs conditioned on given observations. 
Our learning is based on a set of trajectories of the model and observations. 
It aims to correct the pdfs by optimizing likelihood-based loss functions in the sense of the Kullback-Leibler (KL) divergence. Numerical experiments show that we can obtain similar performance when compared with the IEnKF-Q and LETKF methods, 
without the need of localization and inflation techniques.
These numerical results shows the potential advantage of ML based algorithms 
when the used practical algorithms are sub-optimal. 
\section{Introduction}

\subsection{Context}
\justify

In Data Assimilation (DA), the time dependent state
of a system is estimated using two models that are the
\textit{observational model}, which relates the state to physical
observations, and the \textit{dynamical model}, that is used to
propagate the state along the time dimension~\cite{Asch2016}. These models can be written as a Hidden Markov Model (HMM). 

Observational and dynamical models are described using random
variables that account for observation and state errors. Hence DA
algorithms are grounded on a Bayesian approach in which observation
realizations are combined with the above statistical models to obtain
state predictive and posterior density sequences. This estimation is
done in two recursive steps: the \textit{analysis} updates a
predictive density into a posterior one with an incoming observation;
and the \textit{propagation} updates a posterior density into a the
next cycle predictive (or prior) density.

DA methods use additional assumptions or approximations to obtain
closed expressions for the densities so that they can be handled by
computers. Historically in the \textit{Kalman filter} (KF) approach, 
statistical models are assumed to be
Gaussian and \revs{the physical dynamics are assumed to be linear}~\cite{Kalman1960}.
Hence, the propagation and analysis
steps consist in updating \textit{mean and covariance matrix} of \revs{Gaussian}
densities. \revs{A correct estimation of the \textit{covariance matrices}
is crucial since they determine to what extent the predictive density will be corrected
to match observations.}
In the \textit{Ensemble Kalman Filter} (EnKF)
approach, \revs{the covariance matrices} are represented by a
\textit{set of sampling vectors} \revs{to reduce the computational cost of the filter}~\cite{Evensen2009}. 
\revs{When EnKF is used with a small number 
of ensembles, the
covariance matrix estimation becomes low-rank.}
This causes some spurious \revs{correlations} in the covariance matrix
which are filtered by using regularization techniques such as localization and inflation~\cite{Hamilletal01, HoutekamerMitchell01, Asch2016}. EnKF can be used for nonlinear dynamics, however due to the truncation of the statistics up to the second order, in the limit of large ensembles the EnKF filter solution differs from the solution of the Bayesian filter~\cite{Gland11}, except for linear dynamics and Gaussian statistics. Hence, when using these methods for non-linear and non-Gaussian setting there are still open questions in achieving an optimal prediction error in the Bayesian setting.

In this paper, we propose a general supervised learning framework based on Recurrent Neural Network (RNN) for Bayesian DA to approximate a sequence of 
prior and posterior densities conditioned on noisy observations. Section~\ref{sec:DA}
explains the sequential Bayesian DA framework with an emphasis on the \textit{time invariant structure} in the Bayesian DA which is the key property for RNNs. The proposed approach, \textit{Data Assimilation Network} (DAN), is then detailed in Section~\ref{section:DAN} which generalizes both the Elman Neural Network and the Kalman Filter. DAN approximates the prior and posterior densities by minimizing the log-likelihood cost function based on the information loss, related to the cross-entropy. The details of the cost function and the theoretical results for the optimal solution of 
the cost function are presented in Section~\ref{sec:PracticalDAN}.
The practical aspects of the DAN including the architecture and computationally efficient training algorithm are given in Section~\ref{sec:DAN_implementation}. 
We then evaluate the performance of DAN by using fully and partially observed Lorenz-95 system with Gaussian prior and posterior densities in Section~\ref{sec:exp}.
\nrevs{The Lorenz-95 system is non-linear and it is often used as a first-step in meteorology to investigate
potential applications of the proposed method to high-dimensional chaotic systems.}
We compare the performance of DAN with state-of-the-art EnKFs methods, IEnKF-Q and LETKF, in terms of root mean square errors,
and we also provide the stability analysis with respect to the initial condition and the forecast time-interval beyond the training range.
Finally, we provide the conclusions in Section~\ref{sec:Conclusions}.

\subsection{Related work}

With the advances in machine learning and deep learning, 
there has been significant increase in the research of using ML to forecast 
the evolution of physical systems with a data-driven 
approach~\cite{Bruntonetal16, Rudyetal17, raissi2019physics, raissi2017physicsI, raissi2017physicsII, Lietal20, Jiaetal21}. 
Recently, this research has its significant impact on the design of advanced DA algorithms. We next outline three main directions that are related to our research in the hybridization of DA and ML approaches.

In a first direction, one addresses the traditional DA problem
where the goal is to estimate the distribution of a state sequence $x_t$ conditioned on an observation sequence $y_t$, 
by using explicitly an underlying dynamical model $\mcl{M}$. 
\citeA{Harter2012} propose to use Elman Neural Network to learn the analysis equation of KF type algorithm where the dynamics are nonlinear. Their main aim is to reduce the computational complexity without affecting the accuracy. \citeA{McCabe21} focus on the learning of the analysis equation within an EnKF framework. They propose the Amortized Ensemble Filter which aims to improve existing EnKF algorithms by replacing the EnKF analysis equations with a parameterized function in the form of a neural network.

In a second direction, one aims to learn an unknown dynamical model $\mcl{M}$ from noisy observations of $y_t$.
This direction is more ambitious compared to the first one as the dynamics to be learnt can be non-linear or even chaotic.
\citeA{Bocquet2019} propose to use the Bayesian data assimilation 
framework to learn a parametric
$\mcl{M}$ from sequences of observations $y_t$.
The dynamical model is represented by a surrogate model which is formalized as a neural network under locality and homogeneity assumptions.
\citeA{Bocquet2020} extends this framework to the joint estimation of the state $x_t$ and the dynamical model $\mcl{M}$ with a model error represented by a covariance matrix.
They estimate the ensembles of the state by using a traditional Ensemble Kalman Smoother 
based on Gaussian assumption, and then with the given posterior ensemble they minimize for the dynamical model and its error statistics. 
Similarly, \citeA{Brajard2020} propose an iterative algorithm
to learn a neural-network parametric model of $\mcl{M}$. With a fixed $\mcl{M}$,
it estimates the state $x_t$ using the observations $y_t$, and then uses the estimated state
to optimize the parameters of $\mcl{M}$. 
A related work is from
\citeA{Krishnan15}, which introduces a deep KF to 
estimate the mean and the error covariance matrix in KF to model medical data, based on variational autoencoder \cite{Girin21}. 

A third direction, which is what we consider in the present paper, is to estimate the distribution of a state sequence $x_t$ conditioned on a observation sequence $y_t$, without explicitly using the underlying dynamical model $\mcl{M}$ in the propagation. This direction often uses training data in a supervised form of $(x_t,y_t)$. For instance,~\citeA{Fablet2021} propose a joint learning of the NN representation of the model dynamics and of the analysis equation albeit within a traditional variational data assimilation framework. A related work to learn a surrogate model is \citeA{Revach_2022}, 
which proposes a parametric KF to handle partially known model dynamics, replacing explicit covariance matrices by a parametric NN to estimate the model error. \revs{\citeA{Pennyetal22} learns also a surrogate model, based on recurrent neural networks, 
by using only state sequence $(x_t)$ which is then used in a deterministic EnKF framework.}
% the hidden space of

All these approaches consider improving the DA methodologies which are based on an \textit{existing DA algorithm}.  In this work, we
propose a fully data driven approach for Bayesian data assimilation without \textit{relying on any prior DA algorithm} that can be sub-optimal in case of non-Gaussian error statistics and non-linear dynamics.

\subsection{Notation}
We denote a state random variable at time $t$ as $\bx_t$ 
taking their values in some space $\bb X= \bb R^n$ of dimension $n$. 
An observation random variable at time $t$ is denoted by $\by_t$
taking its values in some space $\bb Y$ of dimension $d$ (often $\bb R^d$).
We write a sequence of random variables $\bx_1,\cdots,\bx_t$ as $\bx_{1:t}$. 
A joint probability density of two sequence of random variables $\bx_{1:t}$ and $\by_{1:t}$ with respect to the Lebesgue measure on the finite dimensional 
Euclidean space $\mathbb{X}^t \times \mathbb{Y}^t$ is written as 
$p (x_{1:t},y_{1:t})  = p_{\bx_{1:t},\by_{1:t}}(x_{1:t},y_{1:t})$. 
\revs{We denote the value (realization) of a random variable $\bx$ as $x$}. 
The set of pdfs over $\mathbb{X}$ is denoted by 
$\mathbb{P}_{\mathbb{X}}$.
A conditional pdf for $\bx_t$ \revs{conditioned on the value of $\bs y_t$, i.e.} $\bs y_t = y_t$ is written as 
$p_{\bx_{t} | \by_{t}}(\cdot | y_t)$ $\in \mathbb{P}_{\mathbb{X}}$.
\revs{Given a function $f(x)$ on a measurable space of $\mathbb X$ with measure $p$,
we say $f(x) =0$ $p$-almost everywhere $x$ ($p$-a.e. $x$ in short),
%$p$-a.e. $x$ ($p$-almost everywhere shortly $p$-a.e),
when there exists a measurable set $A$ with $p(A)=0$ such that $f(x)=0$ for all $x \not \in A$. 
}

\section{Sequential Bayesian Data Assimilation}
\label{sec:DA}

In this section, we review the Bayesian optimal solution of sequential 
Bayesian data assimilation for an observed dynamical system
and use its repetitive time-invariant structure to motivate 
the introduction of the DAN framework.

\subsection{Sequential Bayesian Data Assimilation}\label{subsec:sbda}
Data assimilation aims to estimate the state of a dynamical process 
which is modeled by a discrete-time stochastic equation
and observed via available instruments which can be modeled by another stochastic equation~\cite{Asch2016}. These equations are given by the following system:
\begin{subequations}\label{eq:ODSsimu}
  \begin{align}
    \bs x_t & = \mcl{M} \left( \bs x_{t-1} \right) + \bs\eta_t, &\quad \text{(propagation equation)}\label{eq:prop}\\
    \bs y_t & = \mcl{H} \left( \bs x_t \right) + \bs\varepsilon_t, &\quad \text{(observation equation)}\label{eq:obs}
  \end{align}
\end{subequations}
where $\mcl{M} (\cdot)$ is the nonlinear propagation operator that acts on the model state random variable vector at time $t$, $\bs x_t \in \bb X$ and return the model state vector $\bs x_{t+1} \in \bb X$.
$\mcl{H}(\cdot)$ is the nonlinear observation
operator that acts on the state random variable $\bs x_t$ and
approximately returns the observation random variable $\bs y_t \in \bb Y$ at time t. 
Both of these steps may involve errors
and they are represented by 
an \textit{additive model error}, $\bs\eta_t$, and \revs{an \textit{additive observation error}, $\bs\varepsilon_t$}. For example, the observation operator may involve spatial interpolations, physical unit transformations and so on, resulting in measurement errors.
We assume that these stochastic errors are 
distributed according to the pdf $p_{\bs \eta}$ and $p_{\bs \varepsilon}$
and they are i.i.d. along time, independent to the initial state $x_1$. 
Using these assumptions DA problem can be interpreted as a Hidden Markov Model~\cite{Carrassi2018}.

Given such a dynamical model, sequential Bayesian DA aims at quantifying the uncertainty
over the system state each time an observation sample becomes available.
Such an analysis starts by rewriting, under suitable mathematical assumptions, 
the DA system in terms of \textit{conditional probability density functions} 
$p_{\bs x_t | \bs x_{t-1}}(\cdot |x_{t-1}) \in \mathbb{P}_{\mathbb{X}}$
which represents \eqref{eq:prop}, 
and $ p_{\bs y_t| \bs x_t}(\cdot|x_t) \in \mathbb{P}_{\mathbb{Y}}$ 
which represents \eqref{eq:obs}. 
Using these densities, we can quantify the 
uncertainty of the state as a function of the observations. 
This can be done in two steps sequentially using the Bayesian framework: 
the analysis step and the propagation (forecast) step.
Let $p_t^b := p_{\bx_t | \by_{1:t-1}}$ be the \revs{prior} distribution of $\bx_t$ given $\by_{1:t-1}$, 
and $p_t^a := p_{\bx_t | \by_{1:t}}$ be the
posterior distribution of $\bx_t$ given $\by_{1:t}$. 
The \textit{analysis} step computes $p_t^a( \cdot | y_{1:t}) \in \mathbb{P}_{\mathbb{X}} $ from $p_t^b( \cdot | y_{1:t-1}) \in \mathbb{P}_{\mathbb{X}}$ based on Bayes rule, 
\begin{align}
    p_t^a ( \cdot | y_{1:t}) = 
    \frac{ p_{ \by_t| \bx_t }(y_t | \cdot ) \: p_t^b ( \cdot | y_{1:t-1})} {p_{\by_{1:t-1}}(y_{1:t-1})}.
  \label{eq:analysis} 
\end{align}
Here, $p_{ \by_t| \bx_t} (y_t | \cdot )$ is considered as a likelihood function of $x_t$, and
$p_{\by_{1:t-1}}$ is a marginal distribution of observations. Similarly, the \textit{propagation} step computes $p_{t+1}^b(\cdot|y_{1:t} )$ from 
$p_t^a ( \cdot |  y_{1:t} )$, 
\begin{align}
  p_{t+1}^b  (\cdot|y_{1:t} ) 
  &= \int p_{ \bx_{t+1}|\bx_{t}} (\cdot | x) p_t^a ( x |  y_{1:t} )  \mrm d x. 
  \label{eq:propo}
\end{align}
The analysis and forecast steps are then repeated within a given number of cycles (time interval) in which the forecast step provides a prior density for the next cycle. 

Performing the analysis and propagation steps in \eqref{eq:analysis} and \eqref{eq:propo}
with linear dynamics for the propagation operator
$\mcl{M}(\cdot)$ and the observation operator $\mcl{H}(\cdot)$, and using a Gaussian 
assumption for the probabilities $p_{\bs \varepsilon}$ and $p_{\bs \eta}$ reduces to the well known \textit{Kalman filter} (KF, \cite{Kalman1960}). 
The challenge is that the calculation of the pdfs 
become intractable with nonlinear \revs{operators} or non-Gaussian pdfs of the error terms. 
When the dynamics are nonlinear, ensemble type KFs such as Ensemble KF~\cite{Evensen2009} are widely used alternative methods, 
but when used with limited number of ensembles, they require additional \revs{remedies} (see Section \ref{sec:etkfDAN} for further discussions).

\subsection{Time-invariant structure in the Bayesian Data Assimilation}
\label{subsec:invariantBDA}
We review the \textit{invariant structure} of \revs{the Bayesian Data Assimilation (BDA) for the Hidden Markov Model (HMM)} defined in Section \ref{subsec:sbda},
which is a key property to motivate the DAN framework. 
Following the i.i.d. assumptions that we have made on the errors in \eqref{eq:prop} and \eqref{eq:obs}, the conditional pdfs $p_{\bx_{t+1} | \bx_{t}}$ and $p_{\by_{t} | \bx_{t}}$  are time invariant, 
in the sense that for $t=1,2,\ldots$ 
\begin{align*}
 p_{\bx_{t+1} | \bx_{t}} (u | v) & = p_{\bx_{2} | \bx_{1}} (u | v)   \\
 p_{\by_{t} | \bx_{t}} (y | v) & = p_{\by_{1} | \bx_{1}} (y | v)  
\end{align*}
for all $u,v \in \mathbb{X}$ and $y \in \mathbb{Y}$. 

As a result, the conditional pdfs representing the HMM 
are time invariant in the following sense. 
The analysis step \eqref{eq:analysis} 
can then be considered as a \textit{time invariant function}, $a^{BDA}$,
which operates on the prior cpdf, $p_t^b ( \cdot  | y_{1:t-1})  \in \PX$ and
a current observation, $y_t \in \mathbb{Y}$, 
and then return a posterior cpdf
$p_t^a ( \cdot  | y_{1:t} )  \in \PX$:
\begin{align*}
  p_t^a ( \cdot  | y_{1:t} ) = a^{BDA} \left [ p_t^b ( \cdot  | y_{1:t-1}), y_t \right ]. 
\end{align*}
Similarly, according to \eqref{eq:propo}, 
the propagation transformation can be considered as a \textit{time invariant function}, $b^{BDA}$,
that transforms a posterior pdf to a prior pdf, 
$$
p_{t+1}^b ( \cdot  | y_{1:t} ) =  b^{BDA} \left [ p_t^a ( \cdot  | y_{1:t})  \right ].
$$
This presentation of the sequential BDA allows us to see the DA cycle as the composition of \textit{two time invariant} transformations $a^{\mrm{BDA}}$ and
$b^{\mrm{BDA}}$, i.e. each transformation is produced using the \textit{same 
update rule} applied to the previous transformations. 
Exploiting this \textit{repetitive time invariant structure}, 
corresponding to a \textit{chain of events}, leads to a general framework named as the DAN based on recurrent neural networks (RNNs).
We detail these ingredients of the DAN in Section \ref{section:DAN} and Section \ref{sec:DAN_implementation}.

\section{Data Assimilation Networks (DAN)}\label{section:DAN}
In section~\ref{subsection:DANENN} we present DAN, a general framework for DA, which generalizes traditional data assimilation algorithms \revs{such as the Kalman filter and the EnKF} detailed in Section~\ref{sec:kfDAN} and~\ref{sec:etkfDAN}. 
\revs{Thanks to the repetitive structure of BDA, we propose in Section~\ref{sec:PracticalDAN}, 
a log-likelihood cost function based on the information loss to approximate conditional pdfs. Instead of calculating the posterior pdfs analytically, DAN 
aims to learn these pdfs by using sequences of $(x_t,y_t)$ generated from the HMM.
We show theoretically that this framework allows one to handle nonlinear model dynamics and non-Gaussian error distributions
where the Bayesian conditional pdfs are not necessarily Gaussian. 
}

\subsection{DAN framework}
\label{subsection:DANENN}
\begin{figure}[h]
  \centering
	\includegraphics[width=0.8\textwidth]{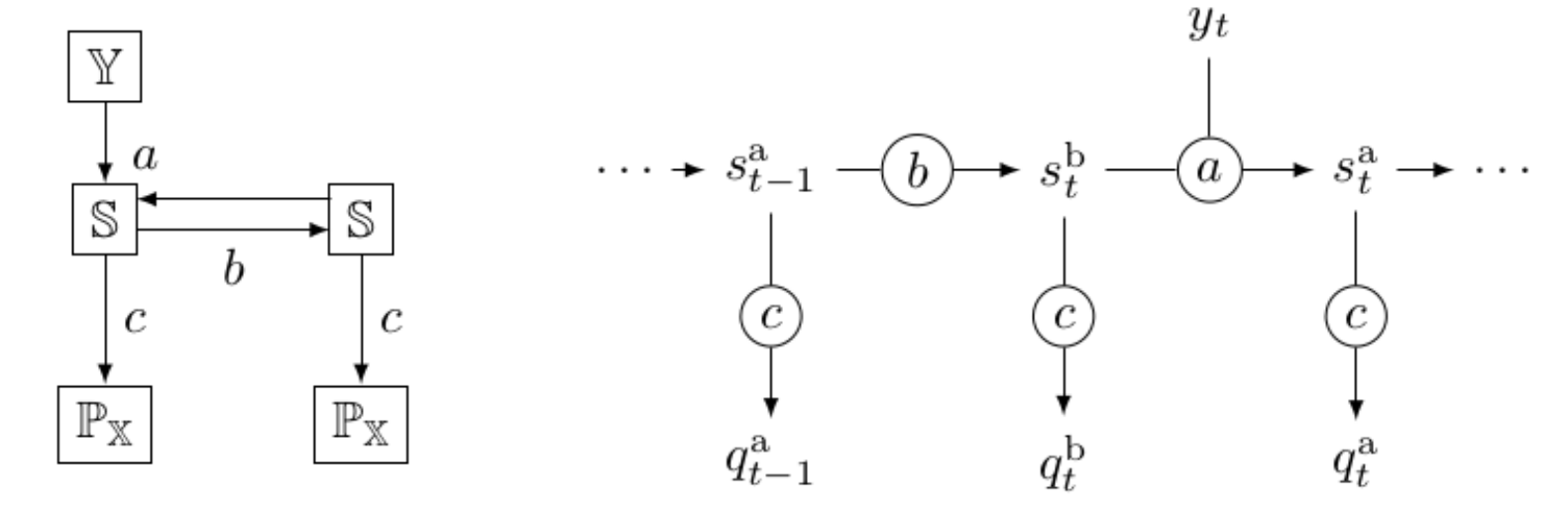}
\caption{Representation of DAN. Left: scheme of DAN. Right: unrolled DAN along time interval.}  
 \label{fig:DAN}
\end{figure}

For a given set $\bb S$, DAN is defined as a triplet of transformations such that 
\begin{subequations}
  \begin{align}
    a &\in \bb S \times \bb Y \rightarrow \bb S,\text{ (analyzer)}\\
    b &\in \bb S \rightarrow \bb S,\text{ (propagator)}\\
    c &\in \bb S \rightarrow \bb P_{\bb X},\text{ (procoder)}
  \end{align}
\end{subequations}
The term ``procoder'' is a contraction of ``probability coder'' 
as the function $c$ transforms an internal representation into an actual pdf over $\bb X$.  A representation of a DAN is given by 
Figure~\ref{fig:DAN}. When $S =  \PX$ and $c$ is identity, this framework encompasses the transformation of $a^{\mrm{BDA}}$ and $b^{\mrm{BDA}}$ in the BDA as a special case. However, it includes also other DA algorithms such as Kalman Filter and Ensemble Kalman Filter.
Such connections are detailed in Section~\ref{sec:kfDAN} and~\ref{sec:etkfDAN}.

One important ingredient of DAN as a general framework for \textit{cycled} DA algorithms is the use of memory to transform prior and posterior densities from one cycle to the next one. In this respect, $\mathbb{S}$ can be interpreted
as a memory space which is a vector space within the DAN framework. 
Considering DAN as a RNN with memory usage naturally make the link with the well-known \textit{Elman Network}. This connection is detailed in Section~\ref{sec:DANEN}.

As a recurrent neural network, we can unroll DAN 
into a sequence of 
transformations.
Given an initial memory $s_0^{\mrm a}\in \bb S_0$, and an observation trajectory
$y_{1:T}\in \bb Y^{T}$, a DAN recursively outputs a predictive and a
posterior sequence such that for $ 1 \leq t  \leq T$,
\begin{align*}
  s_t^b  \coloneqq b\left(s_{t-1}^{\mrm a}\right),\quad&  s_t^{\mrm a} \coloneqq a\left(s_t^{b}, y_t\right)  \\
  q_{t}^{b}   \coloneqq c \left(s_t^b\right),\quad&
  q^{\mrm a}_{t}   \coloneqq c\left(s_t^{\mrm a}\right).
\end{align*}
This recursive application is represented in Figure~\ref{fig:DAN}. Note that $\{q_{t}^b\}_{t=1}^{T}$ and $\{q_t^{\mrm a}\}_{t=1}^{T}$ are
\textit{candidate conditional densities}.
This means that for a given sequence of observations $ y_{1:t} = (y_1 ,\cdots, y_t )$, we have
$q_{t}^b(\cdot | y_{1:t-1}) \in \PX $ and 
$q_{t}^{\mrm a}(\cdot | y_{1:t}) \in \PX $. 
However, these candidate conditional densities are not required to be compatible 
by construction with
a joint-distribution over $\bb X^T \times \bb Y^T$.
As a consequence, we do not assume that there is some 
joint distribution $q(x_{1:T},y_{1:T})$ which induces 
the $q_{t}^b(\cdot | y_{1:t-1})$ and $q_{t}^a(\cdot | y_{1:t})$. 
However, as we shall see in Section \ref{sec:DAN_implementation}, 
the construction of DAN using recurrent neural networks
implicitly imposes some relationships between these candidate conditional densities. 

\subsection{The Kalman Filter as a DAN}
\label{sec:kfDAN}

In the original~\textit{Kalman filter} (KF)~\cite{Kalman1960}, 
both the propagation operator
$ M$ and the observation
operator $H$ are assumed to be affine. 
In this case, the
analysis and propagation transformations preserve Gaussian pdfs that are easily
characterized by their mean and covariance matrix. The analysis and propagation
transformations then simplify to algebraic expressions on these pairs as we
shall see in this section.

\begin{figure}[h]
  \centering
\includegraphics[width=0.8\textwidth]{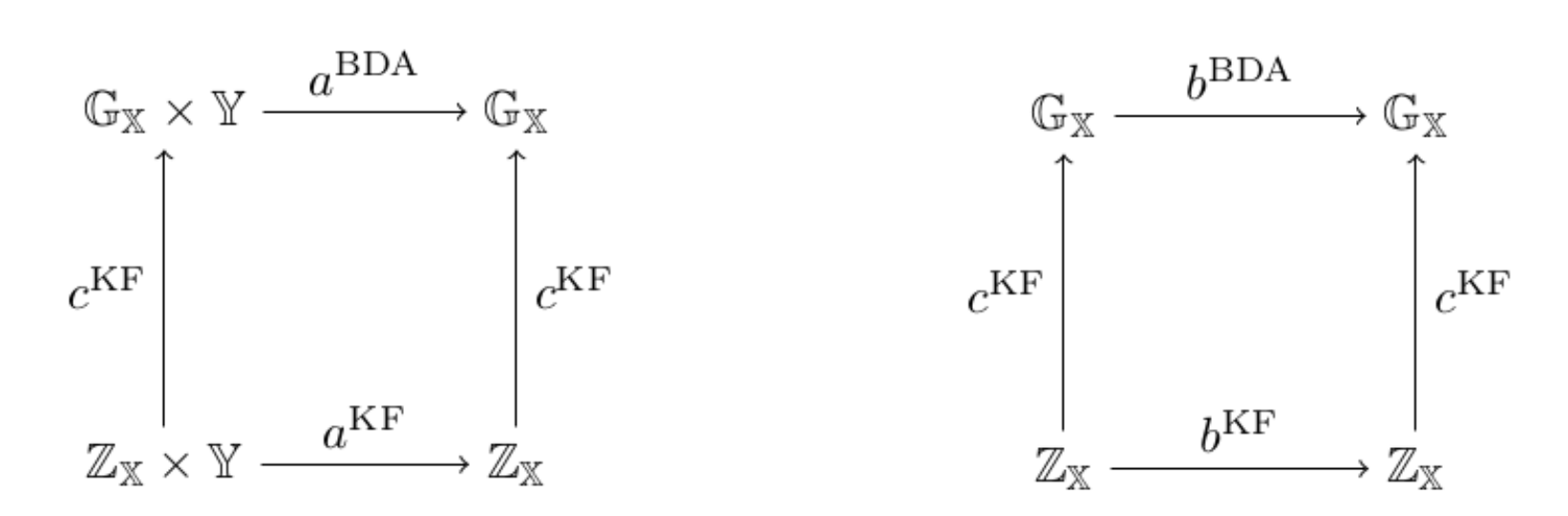}
  \caption{Kalman filter mapping diagram. Left: Commuting diagram for the KF analysis. Right: Commuting diagram for the KF propagation.}
  \label{fig:diagrams}
\end{figure}

Suppose that the internal representation of a Gaussian pdf is formalized by the
injective transformation, $c^{\mrm{KF}} : \bb Z_{\bb{X}} \rightarrow \bb G_{\bb X}$,
\begin{align}
  c^{\mrm{KF}}(s) = \mathcal N \left( \mu,\Sigma \right),\nonumber
\end{align}
where $s \coloneqq (\mu, \Sigma)$, $\mu$ and $\Sigma$ being the mean and covariance matrix respectively and $\bb Z_{\bb X}$ is the set of mean and covariance matrix pairs over $\bb
X$, $\bb G_{\bb X}$ is the set of Gaussian pdfs over $\bb
X$. The KF analysis transformation is the function that transforms such a prior
pair in $\bb Z_{\bb X} $ and an observation $y$ in $\bb Y$ into the posterior pair
in $\bb Z_{\bb X}$, i.e. $a^{\mrm{KF}}: \bb Z_{\bb X} \times \bb Y \rightarrow \bb Z_{\bb X}$, given by
\begin{align}
a^{\mrm{KF}}\left(\mu^{\mrm b},\Sigma^{\mrm b}, y\right) = \left(\mu^{\mrm a}, \Sigma^{\mrm a}\right)
\end{align}
with $\mu^{\mrm a} = \mu^{\mrm b} + \Sigma^{\mrm a}H^{\mrm
  T}R^{-1}\left(y-{H}\left(\mu^{\mrm b}\right) \right)$. \nrevs{When the dimension of the observation $y_t$ is less or equal to the dimension of the state $x_t$, as an alternative we can obtain $\mu^{\mrm a} = \mu^{\mrm b} + K \left(y-{H}\left(\mu^{\mrm b}\right) \right)$ and $\Sigma^{\mrm a}= 
%\left(H^{\mrm T}R^{-1}H + \left(\Sigma^{\mrm b}\right)^{-1}\right)^{-1},
(I - K H)\Sigma^{\mrm b}$ with $K = \Sigma^{\mrm b}H^{\mrm T} (H \Sigma^{\mrm b}H^{\mrm T} + R)^{-1}$.}
  The mapping diagram for the analysis step of the KF is given by the diagram in \Cref{fig:diagrams}, which is a commutative diagram. \revs{We remind that a diagram is said to commute if any two paths between the same nodes compose to give the same map~\cite{pitts_1991}.} 

As well, the KF propagation transformation is the function that transforms a
posterior pair in $\bb Z_{\bb X} $ into the next cycle prior in $\bb Z_{\bb X}$,
i.e. $b^{\mrm{KF}}:\bb Z_{\bb X} \rightarrow \bb Z_{\bb X}$, given by
\begin{align}
    b^{\mrm{KF}} \left(\mu^{\mrm a},\Sigma^{\mrm a}\right) = \left(\mu^{\mrm b}, \Sigma^{\mrm b}\right)
\end{align}
with $\Sigma^{\mrm b}= M\Sigma^{\mrm a}M^{\mrm T} + Q,\,$ $Q$ being the model error covariance matrix and $\mu^{\mrm b} =
M \left(\mu^{\mrm a}\right)$.  The mapping diagram
  for the propagation step of the KF is given by the diagram
in \Cref{fig:diagrams}, which is a commutative diagram.

Unfortunately, \revs{the linearity of $M$ and $H$}
is rarely met in practice and covariance
matrices may not be easy to store and manipulate in the case of large scale
problems. A popular \revs{reduced rank approach} is the \textit{ensemble} Kalman filter that has proven effective in several large scale applications.

\subsection{The Ensemble Kalman Filter as a DAN}\label{sec:etkfDAN}

In the \textit{Ensemble Kalman Filter} (EnKF)~\cite{Evensen2009}, statistics
$\left(\mu,\Sigma\right)\in\bb Z_{\bb X}$ are estimated from an
ensemble matrix $X \in \bb X^m = \bb R^{n\times m}$ having $m$ columns with the
empirical estimators
\begin{subequations}
  \begin{align}
    \mu &= Xu,\\
    \Sigma &= XUX^{\mrm T},
  \end{align}
\end{subequations}
where $u=\left(\frac{1}{m},\ldots,\frac{1}{m}\right)^{\mrm T}\in\bb R^{m},
U=\frac{I_m - m\times uu^{\mrm T}}{m-1}\in\bb R^{m\times m}$ and $I_m\in\bb
R^{m\times m}$ is the identity matrix~\cite{Fillionetal20}. Thus, the algebra over mean and
covariance matrices pairs can be represented by operators on ensembles. In this approach nonlinear operators can be evaluated columnwise on ensembles and
ensembles with few columns may produce low-rank approximations of large scale covariance
matrices. Hence ensembles are an internal representation for the pdfs that are
transformed by the function into a Gaussian pdf, $c^{\mrm EnKF}:\bb X^{m} \rightarrow \bb G_{\bb X}$, 
\begin{align}
    c^{\mrm EnKF}(X) = \mathcal N \left( Xu, XUX^{\mrm T} \right),
\end{align}
when the error covariance matrix $XUX^{\mrm T}$ is full-rank, for instance when $m \ge n$. \revs{In the case when $m < n$, the error covariance matrix become rank deficient resulting in spurious correlations. In this rank-deficient case, we must select a different base measure where the Gaussian distribution is supported,
by using generalized inverse of $XUX^{\mrm T}$ \cite{RAO1973}.}
  
The EnKF analysis transformation is the function that transforms such a prior
ensemble $X_{\mrm b} \in \bb X^m$ and an observation $y \in \bb Y$ into the posterior ensemble $X_{\mrm a} \in \bb X^{m}$, $a^{\mrm{EnKF}}: \bb X^m \times \bb Y \rightarrow \bb X^m$,  given by
\begin{equation}
    a^{\mrm{EnKF}}\left(X_{\mrm b}, y\right) = X_{\mrm a} \quad \text{with}
    \quad X_{\mrm a} = X_{\mrm b} + K\left(Y - Y_{\mrm b} \right)
\end{equation}
where $K=X_{\mrm b}UY_{\mrm{b}}^{\mrm T}\left( Y_{\mrm b}UY_{\mrm{b}}^{\mrm T} +
  R\right)^{-1}\in\bb R^{n\times d}$ is the ensemble Kalman gain, $Y_{\mrm b}=
\mathcal H\left(X_{\mrm b} \right)\in\bb Y^m$ and $Y\in\bb Y^m (= \bb R^{d\times m})$ is a column matrix with $m$ samples
of $\mathcal N\left(y,R\right)$. 

As well, the EnKF propagation transformation is the function that transforms a
posterior ensemble $X_{\mrm a} \in \bb X^m$ into the next cycle prior ensemble $X_{\mrm b} \in \bb X^m$, 
$ b^{\mrm{EnKF}}:\bb X^m \rightarrow \bb X^m$, 
given by
\begin{align}
   b^{\mrm{EnKF}} (X_{\mrm a}) = X_{\mrm b} \quad \text{with} \quad X_{\mrm b} = \mathcal{M}\left(X_{\mrm a} \right) + W
\end{align}
where $W \in \bb X^m$ is a column matrix consisting of $m$ samples distributed according to the Gaussian pdf $\mathcal N\left(0_n,Q\right)$.

In EnKF, as explained above the mean and the covariance matrix for the Gaussian pdf are calculated through ensembles and propagation is performed through the ensembles using nonlinear dynamics. For large-scale nonlinear systems, when one can use only a limited number of ensembles, the error covariance matrix become a rank deficient matrix. This leads to sub-optimal performance~\cite{Asch2016} and may introduce errors during the propagation. For instance, spurious correlations may appear or ensembles may collapse. As a result, for a stable EnKF regularization techniques like localization and inflation needs to be applied~\cite{Hamilletal01, HoutekamerMitchell01,Gharamti18}. Localization consists in filtering out the long-distance spurious correlations in the error covariance matrix. It is not straightforward to find the optimal parameters for the localization, therefore some tuning is required. 
After filtering out these spurious correlations such that the analysis is updated by the local observations, there may be still problem with the use of limited ensembles along the propagation. These small errors may be problematic when they are accumulated through the cycles. This can still lead to filter divergence. A common solution is to inflate the
error covariance matrix by an empirical factor slightly greater than one. The multiplicative inflation compensate errors due to a small size of ensembles 
and the approximate assumption of Gaussian distribution on the error statistics~\cite{Bocquet11}.

%%%%%%%%%%%%%%%%%%%%%%%%%%%%%%%%%%%%%%%%%%%%%%%%%%%%%%%
\subsection{DAN log-likelihood cost function}
\label{sec:PracticalDAN}
%%%%%%%%%%%%%%%%%%%%%%%%%%%%%%%%%%%%%%%%%%%%%%%%%%%%%%%
In this section, we introduce a cost function which allows one 
to optimize the candidate conditional densities, i.e. $q_t^\mrm{a}$ and $q_t^\mrm{b}$,
based on samples of $\bx_{1:T}$ and $\by_{1:T}$. 
The distance between the target conditional densities $p_{t}^b$ 
and $p_{t}^{\mrm a}$ and the candidate conditional densities 
$q_{t}^b$ and $q_{t}^{\mrm a}$
are minimized in the sense of the \textit{information loss}, 
related to \textit{cross-entropy}~\cite{Cover2005}.

\begin{definition}[log-likelihood cost function]\label{def:cost}
  Assume $q = ( q^b_t ,q^{\mrm a}_t )_{t=1}^T \in \bb P = \left(\Pi_{t=1}^{T}\bb Y^{t-1}\rightarrow \bb P_{\bb X} \right)\times  \left(\Pi_{t=1}^{T}\bb Y^{t} \rightarrow \bb P_{\bb X} \right)$
  such that the following log-likelihood cost function is well-defined (i.e. 
  for each $ t \geq 1 $, the Lebesgue integral with respect to $  x_{1:t} $ and $  y_{1:t}$ exists)
  \begin{align}
    \mcl J_t (q_t^b, q_t^{\mrm a}) \coloneqq 
     -\int \left[\ln q^b_{t}(x_t|y_{1:t-1}) + \ln q^{\mrm a}_{t}(x_t|y_{1:t})\right]
    p(x_{1:t},y_{1:t}) \mrm d  x_{1:t} \mrm d y_{1:t}
    \label{eq:Cost_function_t}. 
  \end{align}
  The total log-likelihood cost function is defined as
  \begin{equation}
      \mcl J (q)  \coloneqq \frac{1}{T}\sum_{t=1}^T \mcl J_t (q_t^b, q_t^{\mrm a}).
      \label{eq:totalcost}
  \end{equation}
\end{definition}

The following results show that if $q \in \bb P$, 
the global optima of $ \mcl J$ is 
the Bayesian prior and posterior cpdf trajectories of the HMM. 
\begin{theorem}\label{thm:general}
  Let $\bar{q} \in \arg\min_{ q \in \bb P} \mathcal J (q)$, then
  $\forall t \in \{ 1 , \cdots , T \}$, 
  $\bar{q}_t^b ( x | y_{1:t-1}) = p_t^b  ( x | y_{1:t-1}) $
  for $p_t^b  ( \cdot | y_{1:t-1})$-a.e $x \in \bb X$ and $p$-a.e $ y_{1:t-1} \in \bb Y^{t-1}$.
  Similarly, $\bar{q}_t^{\mrm a} ( x | y_{1:t}) = p_t^{\mrm a}  (x | y_{1:t}) $ 
  for $p_t^{\mrm a} ( \cdot | y_{1:t})$-a.e $x \in \bb X$ and $p$-a.e $ y_{1:t}  \in \bb Y^{t}$.  
\end{theorem}
\begin{proof}
\revs{See \ref{app:thm1}.}
\end{proof}

\revs{Theorem \ref{thm:general} shows that
the objective function of DAN can approximate the Bayesian prior and posterior cpdf
when the candidate pdfs belong to a general functional class (i.e. $q \in \bb P$).
However, the loss function $ \mcl J (q)  $
can not be numerically computed without making the functional class more specific. 
As a common specific case, we next consider 
candidate conditional pdfs as the Gaussian pdfs. }

Let $\bb G_{\bb X}$ be the set 
of Gaussian pdfs over $\bb X$,
and $q  \in \bb G = \left(\Pi_{t=1}^{T}\bb Y^{t-1} 
\rightarrow \bb G_{\bb X} \right)
\times  \left(\Pi_{t=1}^{T}\bb Y^{t} \rightarrow \bb G_{\bb X} \right)$.
For each $J_t(q_t^b,q_t^{\mrm a})$ in Definition \ref{def:cost} to be well-defined,
it is necessary to assume that
the target prior and posterior distributions 
$p_t^b(\cdot |  y_{1:t-1})$ and $p_t^{\mrm a}(\cdot |  y_{1:t})$
have first-order and second-order moments. 
\revs{Under these assumptions, Theorem \ref{thm:gaussian} shows that 
using Gaussian pdfs, one can match the correct mean and covariance of the target prior and posterior cpdf.}

\begin{theorem}\label{thm:gaussian}
  Let $\bar{q} \in \arg\min_{ q \in \bb G} \mathcal J (q)$, then 
  $\forall t \in \{ 1 , \cdots , T \}$, the mean
  and covariance of $\bar{q}_t^b  ( \cdot | y_{1:t-1})$
  equals to the mean and covariance of $p_t^b  ( \cdot | y_{1:t-1})$ for $p$-a.e $ y_{1:t-1} \in \bb Y^{t-1}$.
  Similarly, the mean 
  and covariance of $\bar{q}_t^a  ( \cdot | y_{1:t})$
  equals to the mean and covariance of $p_t^a  ( \cdot | y_{1:t})$ for $p$-a.e $ y_{1:t} \in \bb Y^{t}$.
\end{theorem}
\begin{proof}
\revs{See \ref{app:thm2}.}
\end{proof}

\revs{ 
Theorem \ref{thm:gaussian} indicates that DAN has the capacity to optimally capture non-linear dynamics in terms of first and second-order statistics. Note that here the optimality is defined with respect to the cost function~(\ref{eq:totalcost}). Contrary to KF-based approaches, DAN never uses Gaussian approximations in its internal computations. 
DAN fits the output of the recurrent neural network with Gaussian pdfs.} 

\section{DAN construction and training algorithm}
\label{sec:DAN_implementation}
Having specified the cost function in the previous section, 
we are now going to discuss how to construct 
the components of $a,b,c$ in DAN in order 
to fit training data samples.
To motivate the DAN construction, we first review
its connection with the classical Elman network in Section \ref{sec:DANEN}.
We then specify the construction of a  DAN using recurrent neural networks
in Section \ref{subsec:danrnn}. 
Section \ref{subsec:unrollRNN} and \ref{subsec:algotbptt}
describe how to efficiently train the network. 

\subsection{Connection with Elman network}
\label{sec:DANEN}

\begin{figure}[h]
  \centering
\includegraphics[width=0.9\textwidth]{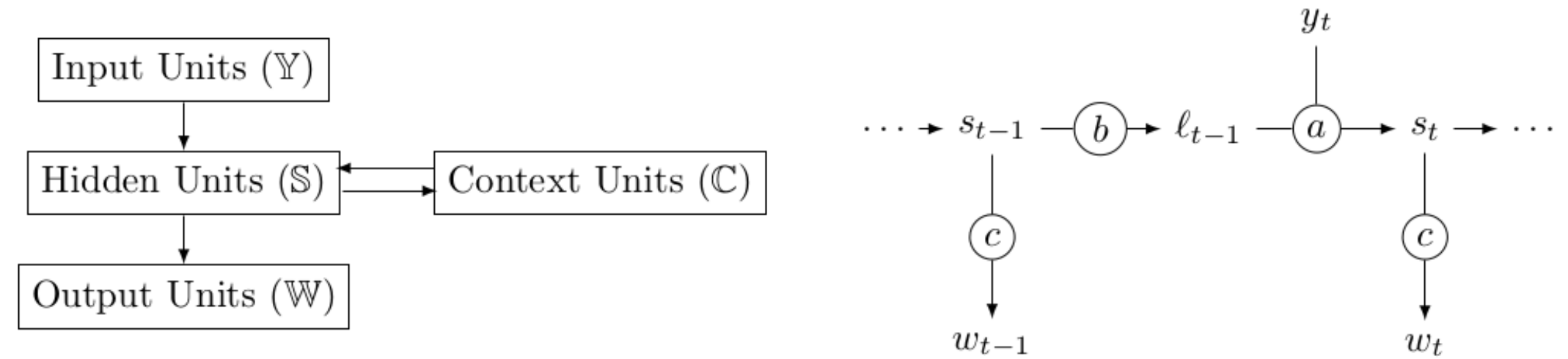}
\caption{Representation of an Elman Network. Left: Scheme of an EN. Right: unrolled EN along time 
interval.}  
\label{fig:EN}
\end{figure}

DAN can be interpreted as an extension of an \textit{Elman network} (EN) \cite{Elman1990} which is a basic structure of recurrent network. 
An Elman network is a three-layer network (input, hidden and output layers) with the addition of a set of context units. These context units provide memory to the network. Both the input units and context units activate the hidden units; the hidden units then feed forward to activate the output units~\cite{Elman1990}. 
A representation of an EN is given in Figure~\ref{fig:EN}.

The context units make the Elman network able to process
variable length sequences of inputs to produce sequences of outputs
as shown in Figure~\ref{fig:EN}. Indeed, given a new 
input $y_t\in\bb Y$ in the input sequence, 
the function $a$ updates
a context memory from $\ell_{t-1} \in \mathbb{C}$ to a hidden state memory 
$s_t= a\left(\ell_{t-1},y_t\right) \in \mathbb{S}$. And the function
$c$ decodes the hidden state memory into an output $w_t = c\left(s_t \right) \in \mathbb{W}$ in the output sequence. The updated hidden state memory is transferred to the context unit via a  function $b$. In a way, the context memory of an Elman network is expected to gather
relevant information from the past inputs to perform satisfactory predictions.
The training process in machine learning
will optimally induce how to manipulate the memory from data.

The similarity between DAN and EN can be made explicit with the analogy 
that the hidden layer is connected to the context units by the function $b$, which includes \textit{time propagation} for DAN. In DAN the hidden unit memory $\mathbb S$ is considered as the same set as the context unit memory $\mathbb C$, and $c$ function decodes both the hidden and the context unit memory into a probability density function.

The EN can not perform DA operations in all its
generality. For instance, EN can not make \textit{predictions} without observations, that is estimating strict future states from past observations. This is because the function $a$ performs both the propagation and the analysis at once. In a way, the EN only produces posterior outputs and no prior outputs while the DAN
produces prior or posterior outputs by applying the procoder $c$ before or after
the propagator $b$ (see Figure~\ref{fig:DAN} and Figure~\ref{fig:EN}). DAN can also produce strict future predictions without
observations by applying the propagator $b$ multiple times before applying the
procoder $c$. Second, the DAN provides a probabilistic representation of the
state i.e. an element in $\bb P_{\bb X}$ instead of an element in $\bb X$. Also,
note that the compositions of $b$ and $c$ make a generalized propagation
operator as it propagates in time probabilistic representations of the state
rather than punctual realizations.

\subsection{Construct DAN using Recurrent Neural networks (RNN)}\label{subsec:danrnn}

We propose to use neural networks to construct 
a parameterized family of DANs. 
Let $\theta$ denote all the weights in neural networks, and 
the memory space
$\bb S $ be a finite-dimensional Euclidean space.
The parametric family of the analyzers and propagators 
are $L$-layer fully connected neural networks:
\begin{subequations}
  \begin{align}
    a_\theta &: \quad  \underbrace{\bb S\times\bb Y \rightarrow \cdots \rightarrow \bb S\times \bb Y}_{ L \; times} \rightarrow \bb S,\\
    b_\theta &: \quad  \underbrace{\bb S\rightarrow \cdots \rightarrow \bb S}_{L \; times},
  \end{align}
\end{subequations}

The construction of $a_\theta$ is built upon $L$ fully-connected layers 
with residual connections. It is based on the LeakyReLU activation function
\cite{Bing2015} 
to improve the trainability when $L$ is large.
For layer $\ell$, the input $v_{\ell-1} \in \bb S\times\bb Y$ 
is transformed into $v_\ell \in \bb S\times\bb Y$ by
\begin{align}\label{eq:ol}
    v_{\ell} = v_{\ell-1} + \alpha_\ell \mrm{LeakyReLU}\left( W_\ell v_{\ell-1}+ \beta_\ell \right) . 
\end{align}
\revs{Taking a vector $v$ as its input, the LeakyReLU function outputs a vector $w$ of the same size. 
For the $i$-th element of $w$, 
$w_i = v_i$ if $v_i \geq 0$; $w_i = a v_i$ if $v_i < 0$, where $a$ is set to $0.01$ by default in our implementation based on Pytorch \cite{NEURIPS2019_9015}}.

An extra linear layer 
is then applied to the output $v_L$ in order to compute a memory state as the output of $a_\theta$. 
The trainable parameters of $a_\theta$ are
$(\alpha_\ell, W_\ell, \beta_\ell)_{\ell \leq L}$ and the weight and bias in the linear layer. 
As illustrated in Figure \ref{fig:DAN},
the input $a_\theta$ at time $t$ 
is a concatenation of $s_t^{b}$ and $y_t$, i.e. $v_0 = (s_t^{b},y_t)$. 
Similarly, $b_\theta$ is constructed from the same $L$ fully-connected layers as in \eqref{eq:ol} by using a different set of trainable parameters. 
The input of $b_\theta$ at time $t$ is set to $s_t^{a}$. 

The procoder $c_\theta$ is specified with respect to the pdf choice of candidate conditional densities. For instance, for the Gaussian case studied in Theorem \ref{thm:gaussian}, $c_\theta$ can be defined as:
\begin{align}\label{eq:procoderGauss}
  c_\theta : \quad  \bb S \rightarrow \bb R^{n + \frac{n\left(n+1\right)}{2}}\rightarrow\bb G_{\bb X}
\end{align}
which is a 
linear layer from $\bb S$ to 
$\bb R^{n + \frac{n\left(n+1\right)}{2}}$, 
followed by a function
that transforms the $n + \frac{n\left(n+1\right)}{2}$ dimensional vector into the mean and the covariance 
of a Gaussian distribution. 
This transformation is detailed in \ref{app:param}. 

\subsection{Training and test loss from unrolled RNN}
\label{subsec:unrollRNN}
In order to train a DAN, we will
unroll the RNN defined by $(a_\theta,b_\theta,c_\theta)$ so as to define the 
training loss computed from $I$ i.i.d trajectories of $(\bx_{1:T},\by_{1:T})$.
We also define the test loss to evaluate the performance of training.

To be clear on how the states $s_t^\mrm a$ and $s_t^b$ 
depend on $a_\theta,b_\theta$
and a given trajectory $y_{1:t}$, we will denote the state (memory) at time $t$ informed by the data up to time $t \shortminus 1$ and generated using a $\theta$-parametric function as $s_{t|t\shortminus1}^{b,\theta}$. Then we can rewrite $s_t^b$ and $s_t^{\mrm a}$ more explicitly as:
\begin{align}
    s_{t|t\shortminus1}^{b, \theta} = b_\theta \left ( s_{t\shortminus1 | t\shortminus1}^{\mrm a, \theta} \right), 
     \quad
    and
    \quad
    s_{t|t}^{\mrm a, \theta} = a_\theta \left ( s_{t|t\shortminus1}^{b, \theta}, y_t \right), 
\end{align}
where $s_{0|0}^{\mrm a, \theta} = s_0$ is an initial memory of RNN independent of $\theta$. 
The procoder $c_\theta$ outputs the pdf
\begin{align}
      q_{t|t\shortminus1}^{b, \theta} ( \cdot | y_{1:t-1} ) = c_\theta \left (  s_{t|t\shortminus1}^{b, \theta} \right ),
    \quad
    and
    \quad
     q_{t|t}^{\mrm a, \theta} ( \cdot |  y_{1:t} ) = c_\theta \left (s_{t|t}^{\mrm a, \theta} \right ).
\end{align}

To define the training loss computed from the $I$ trajectories, 
we introduce a \textit{trajectory-dependent} loss function which 
will be needed to define our online training strategy. 
Let $\left(x^{(i)}_{1:T},y^{(i)}_{1:T}\right)$ be the $i$-th trajectory, 
we write the loss function for the $i$-th trajectory as:
$$
J_t^{(i)} \left ( q_{t|t\shortminus1}^{b, \theta},q_{t|t}^{\mrm a, \theta}\right) = - \log q_{t|t\shortminus1}^{b, \theta} \left( x^{(i)}_t | y_{1:t-1}^{(i)} \right) - \log q_{t|t}^{\mrm a, \theta} \left(x^{(i)}_t | y_{1:t}^{(i)}\right).
$$
The training loss is defined accordingly as a function of $\theta$,
\begin{equation}\label{eq:sampled_theta_cf}
    \frac{1}{T I }  \sum_{t=1}^T  \sum_{i=1}^I J_t^{(i)} \left ( q_{t|t\shortminus1}^{b, \theta},q_{t|t}^{\mrm a, \theta} \right )
\end{equation}

We define the test loss 
$J(\theta)$, as in \eqref{eq:sampled_theta_cf},
by using another $I$ independent trajectories
of $(\bx_{1:T},\by_{1:T})$. 
It allows one to evaluate how well a DAN learns the underlying dynamics of HMM
beyond the training trajectories. 

\subsection{Online training algorithm: TBPTT}\label{subsec:algotbptt}
Direct optimization of the training loss in~(\ref{eq:sampled_theta_cf}) 
is impractical for large-scale problems since to compute the gradient of the loss,
with back-propagation through time, it requires a large computational graph
that consumes a lot of memory ~\cite{Jaeger2002}. 
This limits the training data size $TI$  which, in turn, might
lead to overfitting due to limited data. A workaround is to resort to
gradient descent with truncated backpropagation through time (TBPTT,
\cite{Ronald90, WilliamsZipser95}).
It is commonly used in the machine learning community 
to train recurrent neural networks~\cite{Hao2018, Aicheretal20}.

Starting from $\theta_0$, 
TBPTT is an online method 
which generates a sequence of model parameters 
$\theta_k$ for $k =1,2,\cdots, T$.
\revs{Instead of computing the gradient of the loss \eqref{eq:sampled_theta_cf} with respect $\theta$ which depends on time from $1$ to $T$, 
the idea of TBPTT is to truncate the computation at each iteration $k$ by considering only a part of the gradient from time $k-1$ to $k$.}
Each $\theta_k$ is obtained from $\theta_{k-1}$ 
based on the information of
$I$ training trajectories $\{(x^{(i)}_k,y^{(i)}_k) \}_{i \leq I}$ on-the-fly. 

More precisely, given the initial memories $\{\bar{s}_0^{(i)}\}_{i \leq I}$
and $\theta_0$, 
we update the memory
\[
    \bar{s}_k^{(i)} = a_{\theta_{k-1} } ( b_{\theta_{k-1}} (\bar{s}^{(i)}_{k-1} ), y^{(i)}_{k} ) , \quad k \geq 1
\]
and then we perform the following gradient update, 
\begin{align}\label{eq:tbptt}
    \theta_{k+1} = \theta_{k} - \eta_{k} \frac{1}{I} 
    \sum_{i=1}^I \nabla_\theta J^{(i)}_{k+1} ( 
    c_\theta \cdot b_\theta (\bar{s}_{k}^{(i)}), 
    c_\theta \cdot a_\theta ( b_\theta (\bar{s}_{k}^{(i)} ),  y_{k+1}^{(i)}  )  ) 
    |_{ \theta = \theta_{k} }
\end{align}
where $\eta_{k}$ is the learning rate. \revs{The learning rate is also called the step size in optimization.}
The gradient is computed over the $I$ training trajectories at time $k+1$.
As a result, the optimization is not 
anymore limited in time due to computer memory constraints.

To adjust the learning rate $\eta_{k}$ 
adaptively, we apply
the Adam optimizer \cite{Kingma2014}
to the gradient in \eqref{eq:tbptt}.
This simultaneously 
adjusts the updates of $\theta_k $
based on an average gradient computed from the gradients at previous steps.

\section{Numerical experiments}
\label{sec:exp}
In this section, we present results of DAN 
on the Lorenz-95 system~\cite{Lore95}
using the Gaussian conditional posteriors presented in Theorem \ref{thm:gaussian}. 
We first explain Lorenz dynamics in Section~\ref{sec:Lorenz}, and provide experimental details in Section~\ref{sec:experiment_setup}. Then,
Section \ref{subsec:TBPTT} evaluates the effectiveness of
the online training method TBPTT.
Section \ref{subsec:rmse} compares standard 
rmses performance of DAN to 
state-of-the-art DA methods IEnKF-Q \revs{and LETKF}
using a limited ensemble memory. 
We further study the robustness of DAN 
in terms of its performance on future sequences
beyond the horizon $T$ of the training sequences, 
as well as its sensitivity to the initial 
distribution of each trajectory.

\subsection{The Lorenz-95 system}
\label{sec:Lorenz}
The Lorenz-95 system introduced by~\citeA{Lore95} contains $n$ variables $x_i, i=1,\ldots,n$ and is governed by the $n$ equations:
\begin{equation}
\label{Lorenz_Equation}
	\frac{dx_i}{dt} =-x_{i-2}x_{i-1} + x_{i-1}x_{i+1} - x_i+F.
\end{equation}
In Eq.~(\ref{Lorenz_Equation}) the quadratic terms represent the advection that conserves  the  total  energy,
the linear term represents the damping through which the energy decreases, and the constant term represents
external forcing keeping  the  total  energy  away  from  zero. The $n$ variables may be thought of as values of some atmospheric quantity in $n$ sectors of a latitude circle. 

In this study, we take $n=40$ and $F=8$ which results in some chaotic behaviour. 
The boundary conditions are set to be periodic, 
i.e., $x_0 = x_{40}$, $x_{-1}=x_{39}$ and $x_{41} = x_1$. 
The equations are solved using 
the fourth-order Runge-Kutta scheme, with $\Delta t = 0.05$ (a 6 hour time step).

\subsection{Experiment setup}
\label{sec:experiment_setup}
We study the performance of DAN when trained to map to Gaussian posteriors, i.e. the procoder $c$ function is given by~(\ref{eq:procoderGauss}). 
\revs{This is compared to two  state-of-art baseline  methods of EnKF: Iterative EnKF with additive model error (IEnKF-Q)~\cite{Sakov2018} and Local Ensemble Transform Kalman filter (LETKF)~\cite{Hunt07}.}

A batch of $I$ trajectories of $x \in \mathbb{R}^{40}$ 
is simulated from the resolvant (propagation operator) 
$\mathcal M : \bb R^{40}\rightarrow\bb R^{40}$ 
of the $40$ dimensional Lorenz-95 system. 
To start from a stable regime, 
we use a burning phase which 
propagates an initial batch of states $\{ x^{(i)}_{\mrm{init}} \}_{i \leq I}$
for a fixed number of cycles. The initial states are drawn independently from 
$\mathcal N\left( 3\times 1_{40},I_{40} \right)$. The operator $\mathcal M$ is then applied $10^3$ times (burning time) to the given initial batch of states~{\cite{Sakov2018}}. 
The resulting states are taken as the initial state $x^{(i)}_{1} $.

After the burning phase, 
the Gaussian propagation errors
$\{\eta^i_{t}\}$, sampled independently 
from $\mathcal N\left(0_{40}, 0.01  \times
I_{40} \right)$,
are added to each subsequent propagation 
to get the state trajectories
\begin{align*}
 x^{(i)}_{t+1} &= \mathcal M\left(x^{(i)}_{t} \right) + \eta^{(i)}_{t},
\end{align*}
Then the Gaussian errors $\varepsilon^{(i)}_{t+1}$, sampled independently from
$\mathcal N\left( 0_{40}, I_{40} \right)$, 
are added to the observation
operator evaluations to 
get a training batch of observation trajectories
\begin{align*}
  y^{(i)}_{t+1} &= \mathcal H \left(x^{(i)}_{t+1}\right) + \varepsilon^{(i)}_{t+1}.
\end{align*}
In the numerical experiments
\revs{we consider two cases for the observation network: (1) fully observed, i.e. $\mathcal H$ is taken to be the identity operator $I$, and (2) partially observed, i.e. $\mathcal H$ is taken as a uniform selection operator $H_0$.}
\revs{For any $40$-dimensional vector $x$,
the vector $H_0 x$ preserves half of the grid of $x$, by removing even-indexed elements of $x$.  It is left as a future work to study cases where $H$ is a nonlinear operator.}

\subsubsection{Setup of Baseline}
\label{section_baseline}

\nrevs{
% For comparison, we consider two ETKF methods: 
% IEnKF-Q~\cite{Sakov2018} and LETKF~\cite{Hunt07}. 
The baseline methods, IEnKF-Q and LETKF, 
are implemented with explicit inflation or localization regularization
in order to obtain a good estimation of the covariance matrix of Gaussian densities.
Such regularization is often critical to the final performance of EnKF methods, 
and it often requires the tuning of hyper-parameters whenever the ensemble size $m$ 
is changed~\cite{Asch2016}. 
}

\nrevs{
To illustrate the sensitivity to the hyper-parameter tuning, 
we provide two set of experiments for LETKF: (1) (with case-by-case turning) The filter for each ensemble size is run with the best performance values provided in Table~\ref{fig:infHselection}, found by a 2D grid search for each $m$, named as LETKF$^\ast$. (2) (without case-by-case tuning) The filter for each ensemble is run with the best performance obtained at $m=20$, i.e. the grid search is only run for this $m$ and the obtained optimal hyper-parameters are used for all $m$. We name these experiments simply as LETKF.
}

\nrevs{
We implemented the IEnKF-Q which uses only inflation regularization. 
This allows one to measure the effect of using both inflation and localization regularization in LETKF. 
We present results without case-by-case tuning across different
number of ensembles for EnKF, i.e. $m \in \{5, 10, 20, 30 \}$.
As we do not have localization in IEnKF-Q, 
we fine-tune the inflation hyper-parameter of this method at $m=20$ using grid-search.
We find that on both fully observed and partially observed cases, 
a common inflation parameter 1.1 is close to be optimal among 
$\{ 1.0,1.02,1.03,1.04,1.07,1.08,1.09,1.1,1.2,1.3,2.0 \}$, 
according to the time-averaged posterior (filtering) rmses (see the definition of the rmses in Section \ref{subsec:rmse})
}

\nrevs{
Experiments with the LETKF are performed by using an open source code: DAPPER~\cite[version 1.2.1]{dapper}. 
For each ensemble, we have performed 2D grid search. Localization radius is chosen from the set $\{1,2,4\}$ and the inflation hyper-parameter is chosen from the set $\{1.02, 1.03, 1.04, 1.07, 1.1\}$. 
We also use rotation after the analysis step which is shown to provide better performance for LETKF~\cite{SakovOke08}. 
The inflation and localization radius hyper-parameter values that provide the best performance according to the time-averaged posterior (filtering) rmses are given in Table~\ref{fig:infHselection}. 
}

% \nrevs{The inflation and localization radius values that provide the best performance according to the time-averaged posterior (filtering) rmses are given in Table~\ref{fig:infHselection}. 
% We provide two set of experiments for LETKF: (1) The filter for each ensemble size is run with the best performance values provided in Table~\ref{fig:infHselection}, named as LETKF$^\ast$. (2) The filter for each ensemble is run with the best performance obtained at $m=20$, i.e. the localization radius is set to $4$ and the inflation value is set to $1.04$ for each ensemble. We name these experiments simply as LETKF.}

\begin{table}[h]
  \centering
  \footnotesize
  \nrevs{
  \begin{tabular}{|c|c|c|c|c|c|}
    \hline
    m & 5 & 10 & 20 & 30   \\ 
    \hline
    inflation &  $1.1$  &  $1.07$  & $1.04$ & $1.03$ \\
    \hline    
    local. radius &  $1$  &  $2$  & $4$ & $4$ \\
    \hline
  \end{tabular}
  }
  \footnotesize
  \nrevs{
  \begin{tabular}{|c|c|c|c|c|c|}
    \hline
    m & 5 & 10 & 20 & 30   \\ 
    \hline
    inflation & $1.1$  &  $1.04$ &   $1.03$  &  $1.02$ \\    
    \hline
    local. radius &  $2$  &  $2$  & $4$ & $4$ \\
    \hline
  \end{tabular}
  }
  \caption{\nrevs{Optimal hyper-parameter values of LETKF across various ensemble size $m$ found by 2D grid search. 
  Left: fully-observed case ($H=I$). Right:  partially observed case ($H=H_0$)}.
  }
  \label{fig:infHselection}
\end{table}

\subsubsection{Setup of DAN}
\nrevs{
To make DAN comparable to EnKF in terms of the used memory, 
we set the memory space $\mathbb{S} = \mathbb{R}^{m \times n}$.
Similar to the results without case-by-case turning in LETKF and IEnKF-Q,
hyper-parameters of DAN are only tuned at $m=20$, and 
then fixed across all $m$. 
}
% The functions $a$ and $b$ in the cost function of DAN 
% are constructed by 
% $L=20$ fully connected layers 
% with residual connections (as detailed in Section~\ref{sec:DAN_implementation}).
% To make DAN comparable to EnKF in terms of the used memory, 
% we chose the memory space as $\mathbb{S} = \mathbb{R}^{m \times n}$.

Across \revs{$m \in \{5,10,20,30\}$}, DAN is trained with 
a batch size of $I=1024$ of training samples
for $T=6 \times 10^5$ cycles. 
The initial learning rate $\eta_0$ for the TBPTT is set to be $10^{-4}$.
The initial memory $s_0$ of the RNN is set to be zero, while 
the initial parameter $\theta_0$ of the RNN is mostly set to be random.
More precisely, 
we use the standard random initialization for the weights 
$(W,b)$ of each linear layer implemented in the 
Pytorch software. 
\revs{To train a neural network with a large number of layers $L$, 
we use the ReZero trick \cite{Bachlechner2020}
which sets the initial weight $\alpha_\ell$ in \eqref{eq:ol}
to be zero for each $\ell$.}
\revs{
The functions $a$ and $b$ in the cost function of DAN 
are constructed by 
$L=20$ fully connected layers 
with residual connections (as detailed in Section~\ref{sec:DAN_implementation}).
}

\subsection{Training performance of TBPTT}\label{subsec:TBPTT}

To show the effectiveness of the training method TBPTT specified in \eqref{eq:tbptt}, 
we evaluate the test loss $J(\theta)$ using $I=1024$ i.i.d samples  (defined in Section \ref{subsec:unrollRNN}),
on a sub-sequence of $\theta_{k}$. This allows one to access whether the online method 
is effective to minimize the total loss $\mcl J (q)$ in \eqref{eq:totalcost}.
\revs{The training time of DAN grows with $T$ but it is not sensitive to the choice of $I$. 
This is because our current implementation runs on GPU graphics cards, which allows the computation over $I$ training samples to be in parallel. However, the sequential computation of TBPTT 
can not be done in parallel. One potential improvement of the running time 
is to use a modified version of TBPTT to improve the convergence rate, 
as suggested in \cite[Algorithm 4.2]{chen2022autodifferentiable}. 
}

\begin{figure}[htb]
  \centering \includegraphics[width=0.45\textwidth]{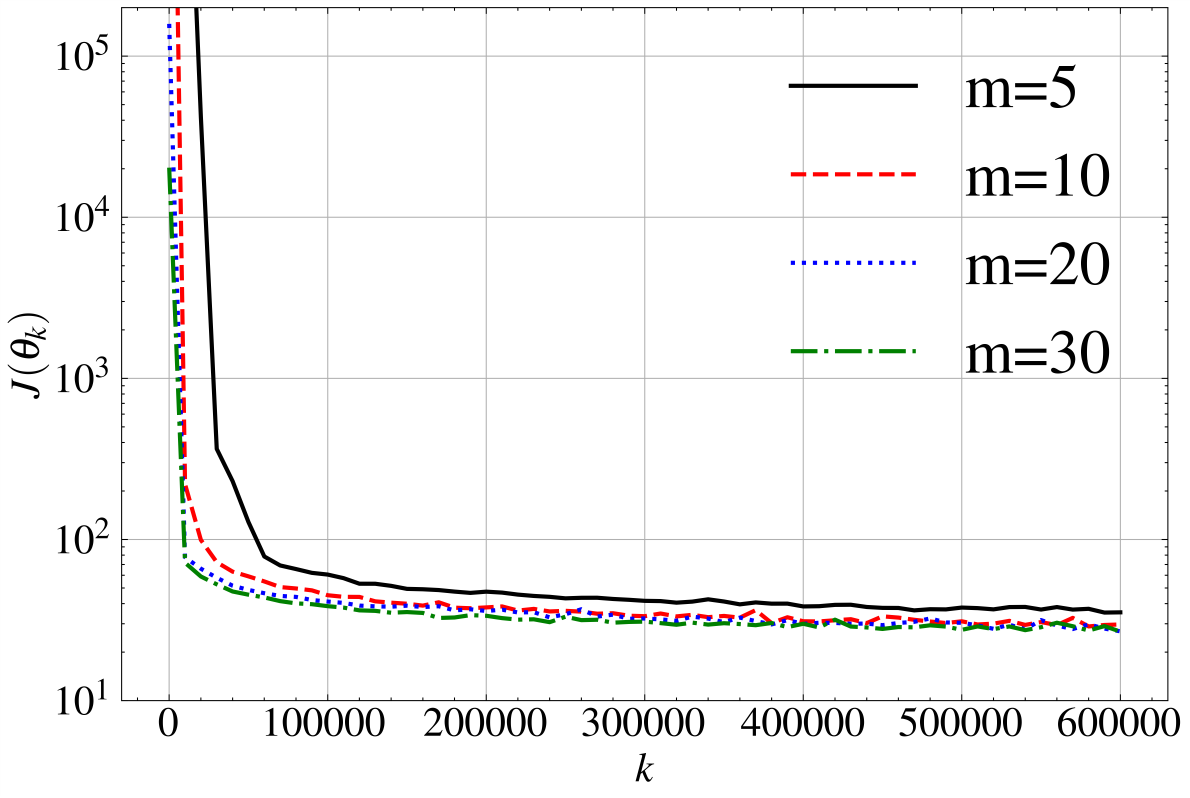}
  \caption{The test loss evaluated at training iterations $\theta_k$ of TBPTT, 
  using various memory size $m$ in DAN. 
  }
  \label{fig:expTBPTT}
\end{figure}

The test loss $J(\theta_{k})$
changes over iteration $k$ are displayed in Figure \ref{fig:expTBPTT}. 
We observe that the minimal loss decreases as $m$ increases, 
suggesting that the performance of DAN is improved with the memory size. 
Moreover, we find that the test loss decreases during the training process, 
which shows that TBPTT implicitly minimizes the test loss $J(\theta)$.
In theory, we expect this to happen for a suitable large memory size $m$ 
because it is proportional to 
the capacity of the neural networks used in DAN: a larger $m$ 
implies a better approximation of the posterior distributions 
due to the universal approximation property of neural networks. 
The trade-off is that a too large $m$ may lead to over-fitting (i.e. a large gap between the training loss and test loss), 
as we use only $I$ finite trajectories of $(x_t,y_t)$ in the training algorithm. 

\subsection{Performance of DAN}\label{subsec:rmse}

After DAN is trained, new observation
trajectories $y_t$ are generated from a new unknown state trajectory $x_t$. 
These testing observations together with a null initial memory vector
are then given as input of the trained DAN 
in a test phase and its outputs are
compared with the unknown state  $x_t$.

To evaluate the accuracy of the trained DAN ($k=T$), 
we compute the accuracy of the
mean $\mu_t^{\mrm a}$ (resp. $\mu_t^b$)
of
$q_{t|t}^{\mrm a,\theta_T} ( \cdot |  y_{1:t} )$ 
(resp. $q_{t|t-1}^{b,\theta_T} ( \cdot |  y_{1:t-1} )$),
evaluated on a test sequence $(x_{1:T},y_{1:T})$.
A standard evaluation in DA is to compute 
rmses, i.e. for $ 1 \leq t \leq T$, 
we compute the following normalized posterior and prior rmses,
\[
    R_t^{\mrm a}   = \frac{1}{ \sqrt{n} } \| x_t - \mu_{t}^{{\mrm a}}  \|,  \quad
    R_t^b  = \frac{1}{ \sqrt{n} } \| x_t - \mu_{t}^{b}  \|
\]

\begin{figure}[h]
  \centering
\includegraphics[width=0.45\textwidth]{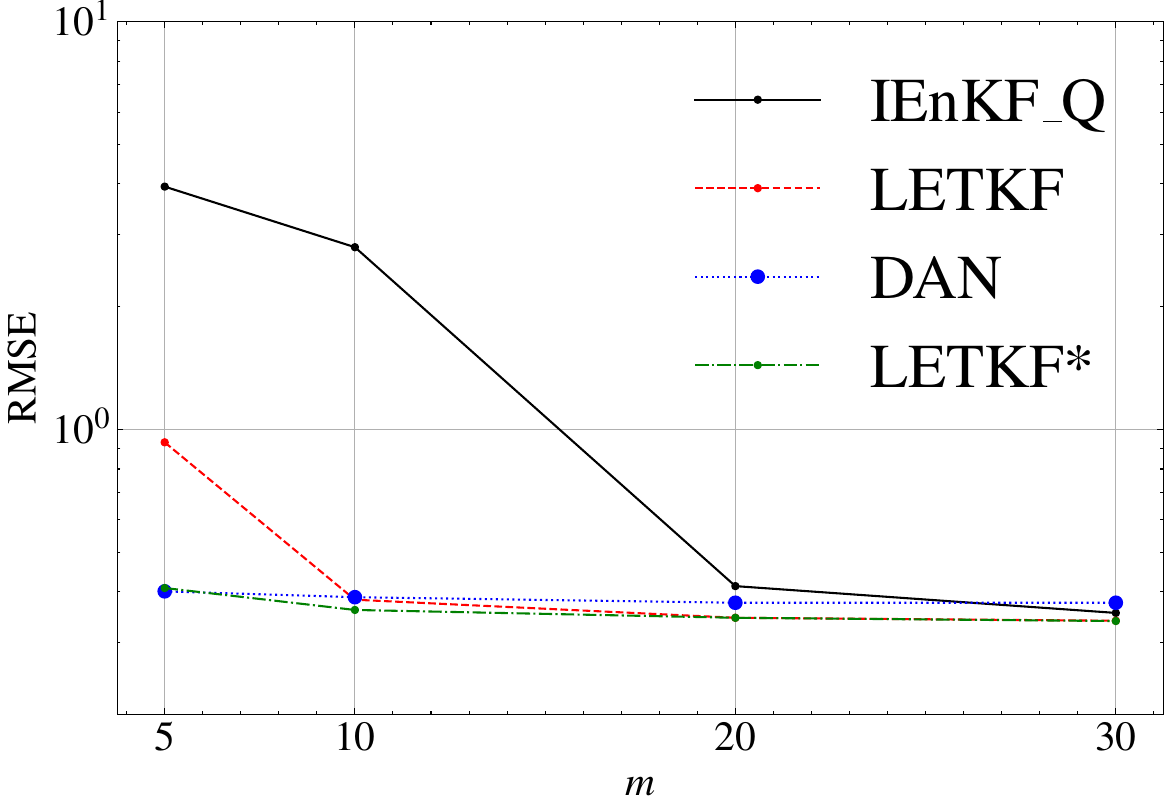}
\includegraphics[width=0.45\textwidth]{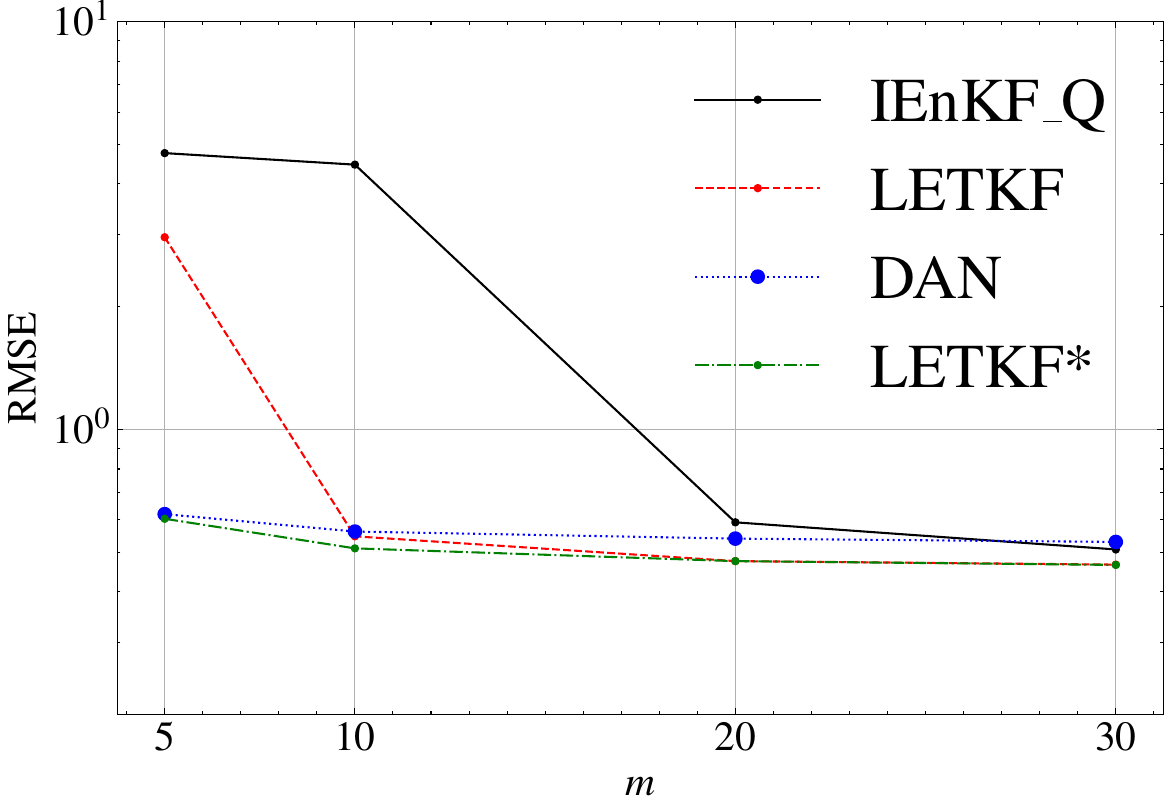}
\caption{Time averaged posterior (filtering) rmses $\frac{1}{T} \sum_{t=1}^T R_t^a$
  using various ensemble size $m$. 
  \revs{Left: fully-observed case ($H=I$). Right:  partially observed case ($H=H_0$).  DAN,   IEnKF-Q and LETKF are tuned at $m=20$; LETKF*  is tuned at each $m$.}}
\label{fig:RMSEa}  
\end{figure}

\begin{figure}[h]
  \centering
\includegraphics[width=0.45\textwidth]{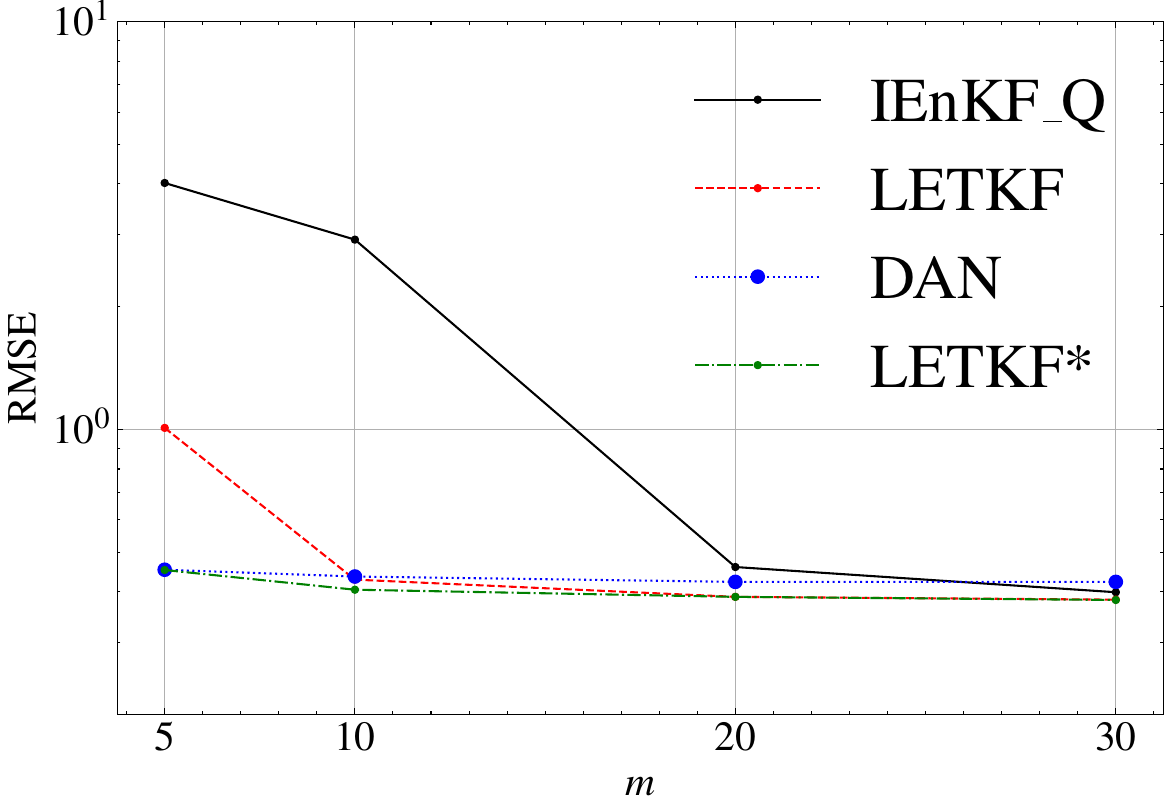}
\includegraphics[width=0.45\textwidth]{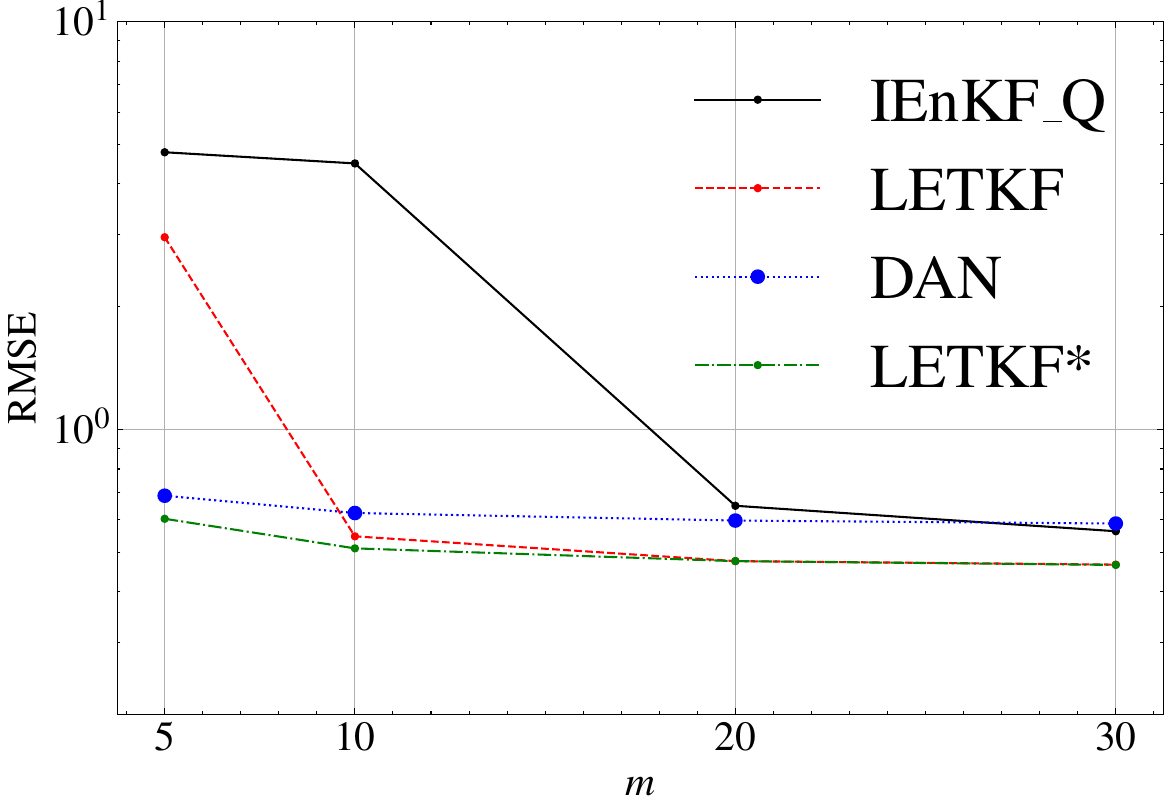}
\caption{Time averaged prior (prediction) rmses $\frac{1}{T} \sum_{t=1}^T R_t^b$
  using various ensemble size $m$. 
  \revs{Left: fully-observed case ($H=I$). Right:  partially observed case ($H=H_0$).  DAN,   IEnKF-Q and LETKF are tuned at $m=20$; LETKF*  is tuned at each $m$.}}
  \label{fig:RMSEb}
\end{figure}

In Figure \ref{fig:RMSEa} and \ref{fig:RMSEb}, 
we compare the averaged rmses of DAN with 
IEnKF-Q and LETKF when the ensemble size $m$ is smaller than the dimension $n$ of the state $x_t$.
For DAN, we report an averaged rmses over $t$, computed at the parameter $\theta_T$ at the last step of training.
These rmses are compared to the two baseline methods, IEnKF-Q and LETKF, over the same range of $t$. 
Recall that we use the same size $m$ to 
define the memory space $\bb S = \mathbb{R}^{m \times n}$ in DAN. 
% \nrevs{We remind that IEnKF-Q is tuned for the inflation parameter at $m=20$ and there is no localization implemented. For LETKF, we run two types of experiments: (1) The filter for each ensemble size is run with the best performance values provided in Table~\ref{fig:infHselection}, named as LETKF$^\ast$. (2) The filter for each ensemble is run with the best performance obtained at $m=20$, i.e. the localization radius is set to $4$ and the inflation value is set to $1.04$ for each ensemble. We name these experiments simply as LETKF.}
% \sixin{Are these reminds necessary? It seems very long.}

Let us first analyze the numerical results for the fully observed case. 
When $m$ is small, IEnKF-Q performs worse 
than DAN, due to sampling errors. 
Note that with the choice $F=8$ in the Lorenz-95 dynamics (Eq.~(\ref{Lorenz_Equation})),
the model has 13 positive and one neutral Lypapunov exponents, i.e. the dimension of the unstable-neutral
subspace is 14~\cite{Trevisanetal2010, BocquetCarrassi2017, Sakov2018, Carrassi2022}.
Therefore, when the model is propagated through
time, small perturbations grow along these directions~\cite{Carrassi2022}. 
This explains why IEnKF-Q does not perform well when $m \le 14$, 
as a result we need to apply localisation and inflation techniques to reduce these sampling errors.
As expected, LETKF and LETKF*, in which localization and inflation techniques are applied with turned parameter values, 
performs much better than the IEnKF-Q. 
DAN performs similarly. When $m=5$, it is slightly better than LETKF* in the fully observed case.
\nrevs{LETKF perfoms much worse than LETKF* and DAN, 
showing how sensitive the method is to the tuning of the inflation and localization hyper-parameters.} 
When $m$ becomes closer to $n$ (e.g. $m=20,30$), we find 
that the posterior and prior rmses of DAN, IEnKF-Q, LETKF and LETKF$^\ast$ are similar, 
with better results for LETKF$^\ast$.
This tendency of rmses as a function the ensemble size $m$ is strongly correlated with the smallest test loss achieved by DAN in Figure \ref{fig:expTBPTT}. 
We observe that for the partially observed case, conclusions are similar as well.
These experiments clearly show that DAN can achieve a comparable performance without using EnKF-type regularization techniques. 

\subsection{Predictive performance and sensitivity to initialization}

As DAN is trained on the time interval $t \leq T$, 
it remains important to evaluate its predictive performance
by considering how well it performs for $t > T$. 
Such performance can be 
measured by the average  
rmses over $T+1 \leq t \leq 2T$ 
instead of over $1 \leq t \leq T$,
evaluated using the trained model parameter ($\theta = \theta_{T}$).
{The posterior rmses for fully observed case are provided in Table \ref{fig:tablePreda}.}
We find that the rmses over $T+1 \leq t \leq 2T$ are close to those over  $1 \leq t \leq T$.
This suggests that DAN has learnt the dynamics of 
the Lorenz system in order to perform well on future trajectories.

\begin{table}[htb]
  \centering
  \footnotesize    
  \begin{tabular}{|l|l|l|l|l|l|}
    \hline
  m & 5 & 10 & 20 & 30  \\ 
    \hline
    DAN & 0.400  &   0.388  & 0.377  & 0.376    \\
    \hline
  \end{tabular}
  \caption{
  Time averaged posterior (filtering) rmses 
  $ \frac{1}{T} \sum_{t=T+1}^{2T} R_t^a$ with various ensemble size.
  }
  \label{fig:tablePreda}
\end{table}

All the earlier results are concerned of the performance of DAN under a fixed burning time.
Using this burning time for the training of DAN, we further evaluate the rmses
on test sequences which have a different burning time. 
It allows us to indirectly access how well recurrent 
structures inherited from the HMM are learnt. 
The results of the ensemble size $m=20$ 
are given in Table \ref{fig:tableBurning}. 
It shows that the performance of DAN 
is not sensitive to the distribution of
the test sample $x_1$ initialized over a wide range of burning time.

We remark that among all the simulations, 
there is always a relatively large error in $R_t^a$ and $R_t^b$ for small $t $
then it decreases very quickly (e.g. $m=20,\mbox{burning}=1000$, both $R_t^a$ and $R_t^b$ get close to a constant level 
when $t \geq 20$).
This transition is needed for DAN to enter a stable regime because the initial memory of the RNN is set to zero.

\begin{table}[htb]
  \centering
  \footnotesize    
  \begin{tabular}{|l|l|l|l|l|l|}
    \hline
    burning time & $10^1$ & $10^3$ & $10^5$ & $10^7$ \\ 
    \hline
    DAN & 0.376  &  0.376  & 0.377   &   0.377  \\
    \hline
  \end{tabular}
  \caption{
  Time averaged posterior (filtering) rmses 
  $ \frac{1}{T} \sum_{t=1}^T R_t^a$ with various burning time at ensemble size $m=20$.
  }
  \label{fig:tableBurning}
\end{table}

\section{Conclusions}
\label{sec:Conclusions}
Based on the key observation that the analysis and propagation
steps of DA consist in applying time-invariant transformations $a$ and $b$ that
update the pdfs using incoming observations, 
we propose a general framework DAN which encompasses well-known state-of the art methods as special cases. 
We have shown that by optimizing suitable likelihood-based objective functions, 
the underlying posterior densities represented by 
these transformations have the capacity to approximate the optimal posterior densities of BDA. 
By representing $a$ and $b$ as neural networks, the estimation problem takes the form
of the minimization of a loss with respect to the parameters of an extended Elman recurrent neural network. 
\revs{As a result, this general framework can be used for nonlinear dynamics and non-Gaussian error statistics.}

\revs{In practice, we need to define the pdfs for the calculation of the loss function. 
As a first step and to be able to compare performance of DAN with the state-of-the-art ensemble methods, we perform numerical experiments with a procoder $c$ which  outputs a Gaussian pdf. Our numerical results on a $40$-dimensional chaotic Lorenz-95 system show that when the ensemble size is small, DAN performs similarly compared to LETKF  which includes regularization techniques such as localization and inflation. For large ensemble size, DAN has similar performance compared to IEnKF-Q and LETKF. It indicates that the DAN framework has the advantage of avoiding some problem-dependent numerical-tuning techniques. }
We also find that DAN is robust in terms of its predictive performance and its initialization.

Although we use a Gaussian approximation of the posterior densities in the procoder $c$, 
it can still happen that the memory space $\bb S$ may encode non-Gaussian information 
of the posterior distributions. 
To analyze why DAN can handle problems with nonlinear dynamics
(even in other nonlinear dynamical systems)
is left as a future study.
\revs{From a practical point of view,  DAN in its current form is not scalable to perform DA when the dimensionality $n$ is very large (e.g. in the order of $10^9$).
To make DAN scalable, different training strategies~\cite{chen2022autodifferentiable,Pennyetal22} will be considered in the future.}

\acknowledgments
This work is partially supported by 3IA Artificial and Natural Intelligence Toulouse Institute, French “Investing for the Future - PIA3” program under the Grant agreement ANR-19-PI3A-0004. \nrevs{The authors wish to thank the two anonymous referees for their constructive comments which helped to improve the manuscript.}

\section*{Open Research}
\noindent All the results and data in this paper can be reproduced from a software 
which is available at 
\url{https://gitlab.com/aniti-data-assimilation/dan_james}.
It can be cited at \url{https://doi.org/10.5281/zenodo.7656199}. 

\bibliography{references}

\appendix

\section{Proof of Theorem \ref{thm:general}}\label{app:thm1}
\begin{proof}
According to \eqref{eq:totalcost}, 
it is sufficient to derive the optimal solution 
of $\mcl J_t(q_t^b,q_t^{\mrm a})$ for each $t$ independently. 
The proof is an application of the KL-divergence~\cite{KL1951} to conditional probability densities.
We re-write $\mcl J_t (q_t^b, q_t^{\mrm a}) $ as 
\begin{align}\label{eq:lienKL}
 \shortminus \int \ln q^b_{t}(x_t|y_{1:t\shortminus1}) p_t^b (x_t | y_{1:t\shortminus1})  p(y_{1:t\shortminus1})  d x_t \mrm d y_{1:t\shortminus1} 
 - \int \ln q^{\mrm a}_{t}(x_t|y_{1:t}) p_t^{\mrm a} (x_t | y_{1:t})  p(y_{1:t})  d x_t \mrm d y_{1:t},
\end{align}
using the property $ p(x_{t},y_{1:t-1})  =  p_t^b (x_t | y_{1:t\shortminus1})  p(y_{1:t\shortminus1})$
and $ p(x_{t},y_{1:t})  =  p_t^a (x_t | y_{1:t})  p(y_{1:t})$.
The first term in \eqref{eq:lienKL} 
can be written as a non-negative conditional relative entropy by including a constant conditional entropy term:
\begin{equation}
\label{eq:KLD}
    \int \left ( \int \ln \frac{ p^b_{t}(x_t|y_{1:t\shortminus1}) }{  q^b_{t}(x_t|y_{1:t\shortminus1})  }  p_t^b (x_t | y_{1:t\shortminus1}) d x_t  \right ) p(y_{1:t\shortminus1}) \mrm d y_{1:t\shortminus1}   \geq 0.
\end{equation}
We have equality in~(\ref{eq:KLD}) if and only if $q^b_{t}( x|y_{1:t\shortminus1}) = p^b_{t}( x |y_{1:t\shortminus1}) $ 
for $p^b_{t}(\cdot | y_{1:t\shortminus1} )$-a.e $x$, and $p$-a.e. $y_{1:t\shortminus1}$
(see a proof in~\cite[Lemma 3.1]{KL1951} and~\cite[Corollary 2.5.4]{Bogachev2007}).
Thus, the minimal solution is given by $\bar{q}_t^b$ as stated 
in the theorem.
Similarly, the minimal solution of the second term \eqref{eq:lienKL} 
is given by the $\bar{q}_t^{\mrm a}$ in the statement.
\end{proof}

\section{Proof of Theorem \ref{thm:gaussian}}\label{app:thm2}

\begin{proof}
We shall only provide a proof for $\bar{q}_t^b  ( \cdot | y_{1:t-1})$
as the proof is similar for $\bar{q}_t^a  ( \cdot | y_{1:t})$. 
Let $\bar{p}^b_{t}( \cdot |y_{1:t-1})$ be the Gaussian distribution 
which has the mean and covariance of  $p^b_{t}( \cdot |y_{1:t-1})$.
Following the proof of Theorem \ref{thm:general}, 
we can rewrite the first term, up to a constant, in \eqref{eq:lienKL} 
into
\begin{equation}\label{eq:qp}
    \int \left ( \int 
      \ln \frac{ \bar{p}^b_{t}(x_t|y_{1:t-1}) }{  q^b_{t}(x_t|y_{1:t-1})  }  
    p_t^b (x_t | y_{1:t-1}) d x_t 
    \right ) p(y_{1:t-1}) \mrm d y_{1:t-1}   
\end{equation}
This is an equivalent minimization problem because we have added a term of $\bar{p}_t^b$ which does not depend on $q_t^b$. 
By definition, $ q^b_{t}(\cdot|y_{1:t-1}) \in \bb G_{\bb X}, \bar{q}^b_{t}( \cdot |y_{1:t-1}) \in \bb G_{\bb X} $, the logarithm term in \eqref{eq:qp} is a quadratic function of $x_t$. As a consequence, we can rewrite \eqref{eq:qp} as
\begin{equation}\label{eq:qp2}
    \int \left ( \int 
      \ln \frac{ \bar{p}^b_{t}(x_t|y_{1:t-1}) }{  q^b_{t}(x_t|y_{1:t-1})  }  
    \bar{p}^b_{t} (x_t | y_{1:t-1}) d x_t 
    \right ) p(y_{1:t-1}) \mrm d y_{1:t-1}   \geq 0 . 
\end{equation}
where we have replaced the density $p^b_{t}$ by $\bar{p}^b_{t}$ because 
they have the same first and second order moments. Note that the inner integral in \eqref{eq:qp2} is the KL divergence 
between $\bar{p}^b_{t}$ and $q_t^b$, so 
its minimal solution $ \bar{q}^b_{t} (\cdot | y_{1:t-1} )$ equals almost surely to $\bar{p}^b_{t} (\cdot | y_{1:t-1} )$.
Therefore the mean and covariance of $\bar{q}^b_{t} (\cdot | y_{1:t-1} )$ and $p^b_{t} (\cdot | y_{1:t-1} )$ match for $p$-a.e. $y_{1:t-1} $. 
\end{proof}

\section{Parameterization of DAN}\label{app:param}

We use the following parameterization of $\mu$ and $\Lambda$ to 
convert the vector $v \in \mathbb{R}^{n + \frac{n (n+1) }{2}  }$ 
in \eqref{eq:procoderGauss} 
into a Gaussian distribution $\mathcal{N}(\mu,\Lambda \Lambda^T)$. 
Let  $v= (v_0, \cdots, v_{n+n(n+1)/2-1})$, we set
\begin{subequations}
  \begin{align}
    \mu &=
          \begin{pmatrix}
            v_0 \\
            \vdots \\
            v_{n-1}
          \end{pmatrix}\in\bb R^n,\\
    \Lambda &=
              \begin{pmatrix}
                e^{v_n} & 0      & \cdots & 0 \\
                v_{2n}    & e^{v_{n+1}} & \ddots & \vdots \\
                \vdots & \ddots  & \ddots & 0 \\
                v_{n+\frac{n\left(n+1\right)}{2}-1} & \cdots & v_{3n-2} &
                e^{v_{2n-1}}
              \end{pmatrix} \in\bb R^{\frac{n(n+1)}{2}}.
  \end{align}
\end{subequations}

The exponential terms in $\Lambda$ ensure 
the positive definiteness of $\Lambda \Lambda^T$. 
This can be easily implemented 
in Pytorch by using the module \texttt{torch.distributions.multivariate\_normal:} 
\texttt{MultivariateNormal(loc=}$\mu$\texttt{,scale\_tril=}$\Lambda$). 

\end{document}